# CONTRIBUTIONS OF NATURAL VENTILATION ON THERMAL PERFORMANCE OF ALTERNATIVE FLOOR PLAN DESIGNS

**Eugénio Rodrigues**[1,2], **Adélio R. Gaspar**[1], **Álvaro Gomes**[2], **Manuel Gameiro da Silva**[1]
[1] ADAI–LAETA, Department of Mechanical Engineering, University of Coimbra, Portugal
[2] INESCC, Department of Electrical and Computer Engineering, University of Coimbra, Portugal

**Abstract**

During the earliest phase of architectural design process, practitioners after analyzing the client's design program, legal requirements, topographic constraints, and preferences synthetize these requirements into architectural floor plan drawings. Design decisions taken in this phase may significantly contribute to the building performance. On account of this reason, it is important to estimate and compare alternative solutions, when it is still manageable to change the building design. For instance, issues like how spaces are arranged in the floor plan, how exterior openings are oriented and dimensioned, and how the floor plan is oriented result in airflow paths that have considerable impact to the spaces' indoor thermal comfort.

The authors have been developing a prototype tool to assist architects during this initial design phase. It is made up of two algorithms. The first algorithm generates alternative floor plans according to the architect's preferences and requirements, and the client's design program. It consists in one evolutionary strategy approach enhanced with local search technique to allocate rooms on several levels in the two-dimensional space. The second algorithm evaluates, ranks, and optimizes those floor plans according to thermal performance criteria. The prototype tool is coupled with dynamic simulation program, which estimates the thermal behavior of each solution. A sequential variable optimization is used to change several geometric values of different architectural elements in the floor plans to explore the improvement potential.

In the present communication, the two algorithms are used in an iterative process to generate and optimize the thermal performance of alternative floor plans. In the building simulation specifications of EnergyPlus program, the airflow network model has been used in order to adequately model the air infiltration and the airflows through indoor spaces. A case study of a single-family house with three rooms in a single level is presented. Eight alternative floor plan design solutions were automatically generated with different shape and space arrangements. The solutions were then evaluated and iteratively optimized according to thermal performance criteria. The location, weather data, and constructive system are kept the same for all design solutions. The results of the contribution of natural ventilation to thermal performance are ranked and compared in a four scenarios analysis.

## 1   Introduction

Design decisions taken in early architectural design process phase may significantly contribute to the final building performance. For instance, how spaces are arranged in the floor plan, which orientation has the floor plan, what size and orientation of each exterior opening, and how these are controlled by the occupants, result in airflow paths that may have significant contribution to the spaces' indoor thermal comfort.

The developed prototype tool to assist architects during the initial design phase of space planning is made up of two algorithms. The first algorithm generates alternative floor plans according to the architect's preferences and requirements, and the client's design program (Rodrigues, Gaspar, &

Gomes, 2013a, 2013b, 2013c). It consists in a hybrid evolutionary strategy approach enhanced with local search technique to allocate rooms on one (Rodrigues et al., 2013b, 2013c) or several levels (Rodrigues et al., 2013a) in n-layered two-dimensional plane. It is named Evolutionary Program for the Space Allocation Problem (EPSAP). The second algorithm evaluates, ranks, and optimizes those generated floor plans according to thermal performance criteria (Rodrigues, Gaspar, & Gomes, 2014a, 2014b). This algorithm, named Floor plan Performance Optimization Program (FPOP), is coupled to the dynamic simulation program EnergyPlus (v8.1.0), which estimates the thermal behavior of each floor plan solution. A sequential variable optimization is used to change several geometric values of different architectural elements in the floor plans to explore the improvement potential.

Natural ventilation has a significant contribution on the thermal comfort, energy use, and air quality of a building (Faggianelli, Brun, Wurtz, & Muselli, 2014). Different ventilation strategies have been studied to improve building performance in different sectors such as residential (Firląg & Zawada, 2013), commercial (Carrilho da Graça, Martins, & Horta, 2012; Ng, Musser, Persily, & Emmerich, 2013), and industrial (Huang & Lin, 2014). Different aspects influence the airflow paths in buildings, like the climatic conditions (Faggianelli et al., 2014), the building's shape, the interior space arrangement (Huang & Lin, 2014; Kao, Chang, Hsieh, Wang, & Hsieh, 2009), the opening's design (Stavrakakis, Zervas, Sarimveis, & Markatos, 2012), and the occupants' venting behavior (Karava, Stathopoulos, & Athienitis, 2011). This paper first presents eight alternative floor plan designs for a single-family house located in Porto, Portugal, that were generated, assessed, ranked, and optimized according to thermal performance criteria. This set of floor plan solutions is used as reference to compare four natural ventilation scenarios. Each scenario varies according to the occupants' behavior in opening the windows and doors to create airflow paths, thus contributing to reducing the building indoor air temperature.

## 2   Methodology

The methodology consists in generating a set of alternative floor plans, to optimize those floor plans from thermal performance point-of-view having the openings closed (only air infiltration through cracks are considered), to perform thermal assessment of four naturally ventilated design scenarios, and to compare those results.

The floor plan designs are generated using the EPSAP algorithm (Rodrigues et al., 2013a, 2013b, 2013c). The algorithm is a hybrid evolutionary technique that couples an Evolutionary Strategy (ES) with a local search technique. EPSAP is capable of generating multi-level floor plans where vertical circulation spaces are treated as flexible and parametric objects that evolve along the search process. The algorithm generates set of alternative solutions according to the user's preferences and requirements, which may be topological (connectivity between spaces, openings orientation, and preferable location for specific spaces) and geometric (minimum floor areas, minimum window widths, maximum construction area, etc.). In the current case, EPSAP is used to generate eight alternative single-family houses. The design program has a hall, a corridor, a kitchen, a living room, three bedrooms, and two bathrooms (see Figure 1).

After the floor plans have been generated, their geometry is optimized using FPOP algorithm from thermal performance criteria (Rodrigues et al., 2014a, 2014b). The algorithm consists in a sequential variable optimization procedure, where different building geometry variables are changed with the aim of minimizing the thermal penalties obtained from the cost function Eq. (1).

$$f_{th}(I) = \sum_{i=1}^{N_d} \sum_{t=1}^{N_t} \left( f_{df}\left(T_i(t), T_1, T_2\right) \times f_{oc}(i,t) \right) \qquad (1)$$

The geometric variables are the floor plan orientation and reflection, openings position and size, interior wall position, and size of overhangs and fins. The cost function determines degree-hours of discomfort in each space of the floor plan and multiplies it by a corresponding weight ($f_{df}$, Eq. (2)). Then, the result is multiplied with an occupancy factor ($f_{oc}$, Eq. (3)).

$$f_{df}(T, T_1, T_2) = \begin{cases} w_1(T_1 - T) & \text{if } T_1 > T \\ w_2(T - T_2) & \text{if } T > T_2 \\ 0 & \text{otherwise} \end{cases} \quad (2)$$

$$f_{oc}(i, t) = \begin{cases} o & \text{if space } i \text{ occupied at time } t \\ v & \text{otherwise} \end{cases} \quad (3)$$

This allows the user to specify his preferences on heating and cooling needs, and to specify his confidence on occupancy schedules (for more details see Rodrigues et al., 2014b). The thermal comfort limits ($T_1$ and $T_2$) are determined according to EN 15251:2007 for operative temperatures for buildings without mechanical cooling and heating systems. The value for $w_1$, $w_2$, and $o$ was 1, and for $v$ was 0.3.

The thermal performance of each solution is estimated using a dynamic simulation program (EnergyPlus). FPOP has stored in its database the constructive system, with materials physical properties, which is in accordance with the Portuguese building regulation (see Table 1). Also in the same database, the simulation specifications are set for each space internal gains (equipment, artificial lighting, occupants), schedules of use (see Table 2 and Table 3), weather data, and location information (latitude, longitude, altitude for the city of Porto, Portugal). The airflow network model of EnergyPlus is used to simulate the air infiltration and natural ventilation. The openings and surfaces cracks parameters are set to have approximately 0.4 air changes per hour (ACH) on average. During the optimization procedure, the openings are set close.

In the end of the FPOP optimization, four natural ventilation scenarios are carried out. The openings are only opened if a space is occupied (see spaces occupancy in Table 2), the indoor air temperature is equal or above 22ºC, and if the outdoor air temperature is below indoor air temperature ($d_1$ and $d_2$, lower limit of temperature difference to begin to open the opening and upper limit of temperature difference to have fully open the opening). In each scenario, the windows and doors are modulated to simulate the occupants' behavior in controlling the natural ventilation. Because spaces have different occupancy schedules, is expected to have different airflow paths during the day, between scenarios, and in each floor plan. The scenarios A to D represent the decrease of reaction of the occupants to overheated environment. In the first scenario (A), the openings are immediately open as soon $d_1$ is 0ºC and it is fully open $d_2$ reaches 6ºC. Similarly scenario B has $d_1$ and $d_2$ equal to 1ºC and 4ºC, scenario C $d_1$ and $d_2$ equals to 2ºC and 8ºC, and scenario D has $d_1$ and $d_2$ equal to 6ºC and 12ºC, respectively.

Finally, the scenarios are compared and analyzed for each of the floor plans to determine the contribution of natural ventilation.

**Table 1:** *Reference U-values and materials physical properties*

| Element | U-value | Layer | T{cm} | C{W/m-K} | D{kg/m3} | SP{J/kg-K} | TA | SA | VA |
|---|---|---|---|---|---|---|---|---|---|
| Ceiling/Slab | 2.60 | High weight concrete | 20.0 | 1.73 | 2242.60 | 836.80 | 0.90 | 0.65 | 0.65 |
| | | Hardwood | 3.00 | 0.20 | 825.00 | 2385.00 | 0.90 | 0.78 | 0.78 |
| Exterior door | 2.86 | Insulation | 1.00 | 0.04 | 32.00 | 836.80 | 0.90 | 0.50 | 0.50 |
| | | Hardwood | 1.00 | 0.20 | 825.00 | 2385.00 | 0.90 | 0.78 | 0.78 |
| | | Plaster | 2.00 | 0.43 | 1250.00 | 1088.00 | 0.90 | 0.60 | 0.60 |
| | | Dense brick | 11.00 | 1.25 | 2082.40 | 920.50 | 0.90 | 0.93 | 0.93 |
| Exterior wall | 0.43 | Insulation | 8.00 | 0.04 | 32.00 | 836.80 | 0.90 | 0.50 | 0.50 |
| | | Concrete block | 15.00 | 1.73 | 2242.60 | 836.80 | 0.90 | 0.65 | 0.65 |
| | | Plaster (gypsum) | 2.00 | 0.22 | 950.00 | 840.00 | 0.90 | 0.60 | 0.60 |
| | | High weight concrete | 20.00 | 1.73 | 2242.60 | 836.80 | 0.90 | 0.65 | 0.65 |
| Floor | 0.45 | Insulation | 8.00 | 0.04 | 32.00 | 836.80 | 0.90 | 0.50 | 0.50 |
| | | Lime plaster | 2.00 | 0.80 | 1600.00 | 840.00 | 0.90 | 0.50 | 0.50 |
| | | Hardwood | 1.50 | 0.20 | 825.00 | 2385.00 | 0.90 | 0.78 | 0.78 |
| | | Hardwood | 0.50 | 0.16 | 720.80 | 1255.20 | 0.90 | 0.78 | 0.78 |
| Interior door | 1.36 | Chipboard | 3.00 | 0.07 | 430.00 | 1260.00 | 0.90 | 0.78 | 0.78 |
| | | Hardwood | 0.50 | 0.16 | 720.80 | 1255.20 | 0.90 | 0.78 | 0.78 |
| | | Plaster (gypsum) | 2.00 | 0.22 | 950.00 | 840.00 | 0.90 | 0.60 | 0.60 |
| Interior wall | 2.17 | Concrete block | 7.00 | 1.73 | 2242.60 | 836.80 | 0.90 | 0.65 | 0.65 |
| | | Plaster (gypsum) | 2.00 | 0.22 | 950.00 | 840.00 | 0.90 | 0.60 | 0.60 |
| | | Slag | 1.50 | 1.44 | 881.00 | 1673.60 | 0.90 | 0.55 | 0.55 |
| | | Felt and membrane | 1.00 | 0.19 | 1121.30 | 1673.60 | 0.90 | 0.75 | 0.75 |
| Roof | 0.37 | Dense insulation | 10.00 | 0.04 | 91.30 | 836.80 | 0.90 | 0.50 | 0.50 |
| | | High weight concrete | 20.00 | 1.73 | 2242.60 | 836.80 | 0.90 | 0.65 | 0.65 |
| | | Plaster (gypsum) | 2.00 | 0.22 | 950.00 | 840.00 | 0.90 | 0.60 | 0.60 |
| **Element** | **U-value** | **Type** | **g-value** | **VT** | | | | | |
| Window | 2.60 | Double Glazed Win. | 0.63 | 0.70 | | | | | |

T - layer thickness; C - conductivity; D - density; SH - specific heat;
TA - thermal absorptance; SA - solar absorptance; VA - visible absorptance; and

**Table 2:** *People occupancy, activity, and schedule by space*

| Space | Time (1–24) | {W} | Pp |
|---|---|---|---|
| Hall | | 190 | 2 |
| Kitchen | | 190 | 2 |
| Living room | | 110 | 5 |
| Bathroom | | 207 | 1 |
| Corridor | | 190 | 2 |
| Bedroom | | 72 | 2 |
| Bedroom | | 72 | 2 |
| Bedroom | | 72 | 1 |
| P. Bathroom | | 207 | 1 |

**Table 3:** *Light/equipment gains and schedule by space*

| Space | Time (1–24) | {W/m2} |
|---|---|---|
| Hall | | 7.0 |
| Kitchen | | 10.0 |
| Living room | | 10.0 |
| Bathroom | | 7.0 |
| Corridor | | 7.0 |
| Bedroom | | 7.0 |
| Bedroom | | 7.0 |
| Bedroom | | 7.0 |
| P. Bathroom | | 7.0 |

## 3   Results and discussion

The generated and optimized floor plans are presented in Figure 1. These are sorted according to thermal performance carried out using Eq. (1). Openings were set close all time during this procedure. On the top-right corner of each floor plan is represented the North orientation, in the top-left is indicated the design number, the EPSAP design penalties (all solutions satisfy all preferences and requirements of the user), and the FPOP thermal penalties. It is possible to observe the diversity of floor plans in their shape and interior configuration. Windows facing South, East, and West tend to have overhangs and fins (dashed lines) to minimize overheating. This scenario thermal performance results are summarized in Table 4, column NoVent. In the same table, the results for the remaining scenarios are also presented (A to D) with indication of the best improvement percentage for each design. It is noticeable that the improvement percentage due to natural ventilation may vary between 4.7% and 7.9%. In bold are marked the best performance of all designs and in italic are marked the best scenarios for each design. As shown, the design 120 has the best performance in all natural ventilation scenarios.

**Table 4:** *Design thermal penalties for each natural ventilation scenario*

| Design | Scenarios | | | | | Imp {%} |
|---|---|---|---|---|---|---|
| | A | B * | C * | D | NoVent * | |
| 120 | **19681.5** | **20191.3** | *19656.0* | **20148.5** | **21216.0** | 7.4% |
| 125 | *21166.0* | 21675.2 | 21172.6 | 21425.7 | 22940.7 | 7.7% |
| 116 | *26787.5* | 27322.0 | 26805.1 | 27352.7 | 28627.4 | 6.4% |
| 114 | 27390.4 | 27848.3 | *27367.8* | 27428.7 | 28887.7 | 5.3% |
| 115 | 27654.8 | 28017.9 | *27653.8* | 27708.2 | 29033.2 | 4.8% |
| 127 | 27324.7 | 28089.5 | *27140.4* | 27484.8 | 29473.3 | 7.9% |
| 117 | 28759.9 | 29206.3 | *28728.6* | 28785.2 | 30144.9 | 4.7% |
| 119 | 30215.8 | 30987.6 | *30150.3* | 30212.3 | 32495.9 | 7.2% |
| **Avg** | 26122.6 | 26667.3 | *26084.3* | 26318.3 | 27852.4 | |

\* Scenarios compared in Figure 2
Imp - percentage of improvement of the best scenario in comparison to reference scenario;
Avg - scenario penalties average; NoVent – Reference scenario with no ventilation.

Contrary to what would be expected, scenarios where openings are open with smaller $d_1$ and $d_2$, are not the ones which perform best. In fact, the best scenario was C, which may infer a compromise between under- and over-heating penalties. Figure 2 depicts the difference of relative penalties per day for whole year between the reference scenario NoVent (horizontal dashed line) and scenarios B (strong line) and C (thin line). As it is possible to observe and despite scenario B being more reactive in opening the windows and doors, every floor plan have lesser reduction of penalties than in scenario C. In fact, in the scenario B the penalties never increased, in comparison to reference scenario, however in the case of scenario C, and despite penalties increased in some periods (small positive picks around the year), the benefits in the summer period are greater. As expected, the period of time that floor plans have the greatest benefit from natural ventilation is from May to October.

## 4   Conclusion

Natural ventilation has an important contribution to the spaces thermal performance. As shown, alternative floor plan designs, which satisfy the same user's requirements and preferences, have different benefits depending on the occupants' behavior in using natural ventilation to cool their homes. However, it is possible to conclude that judicious use of natural ventilation must be taken, even when guarantee that openings are only open during thermal comfort periods, as a small tolerance over higher thermal comfort limit may result bigger benefits in a whole year evaluation.

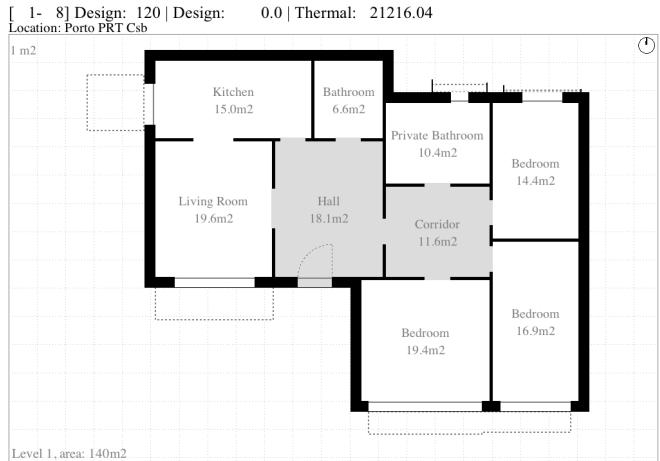
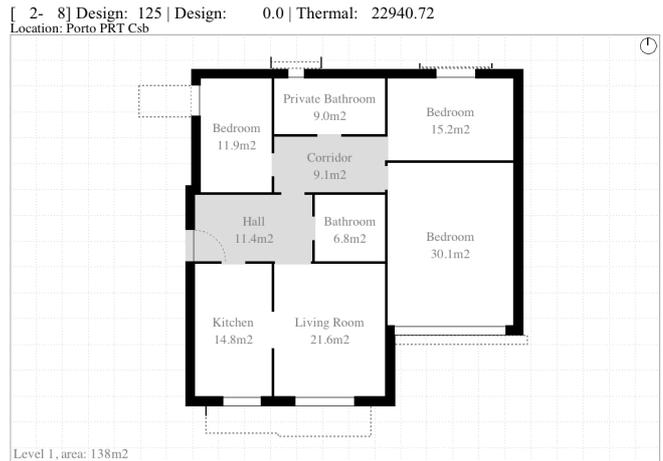
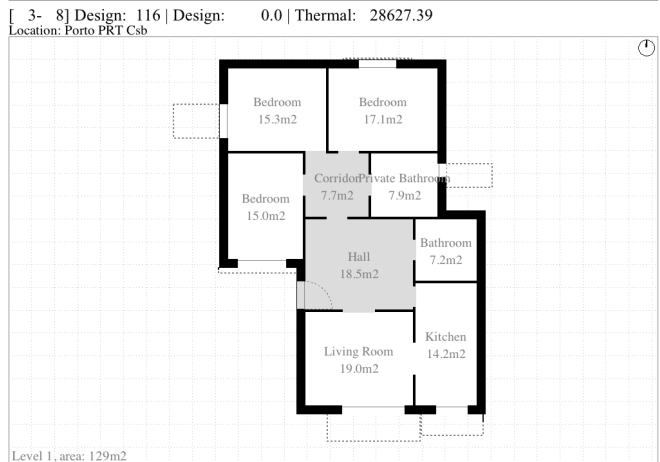
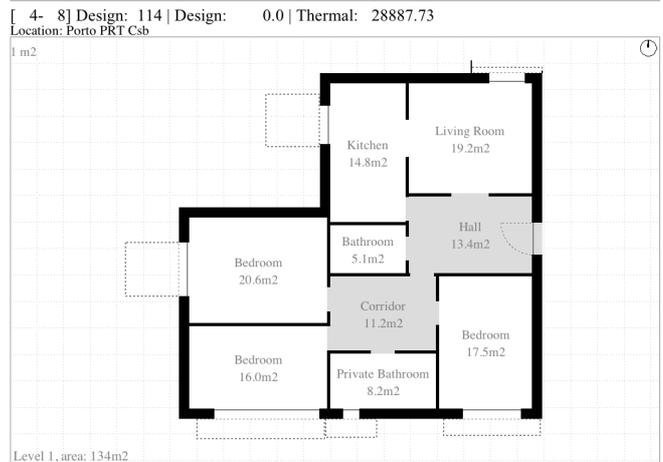
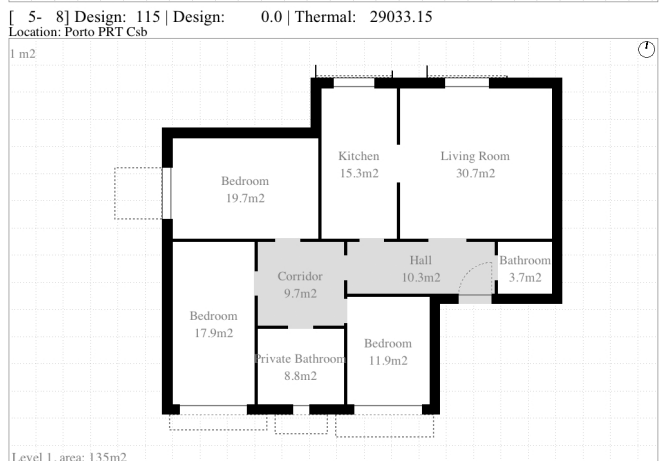
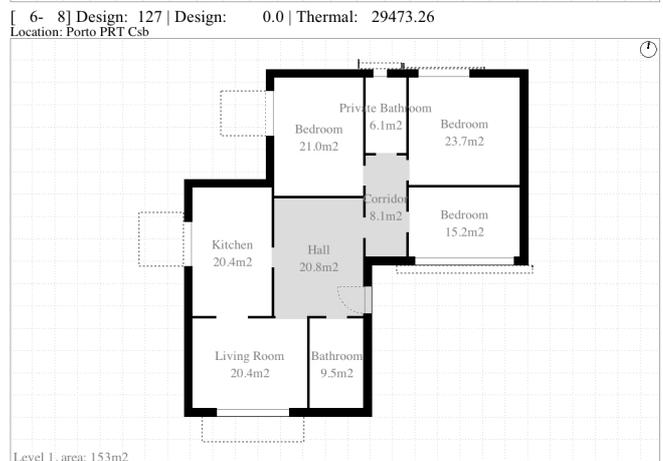
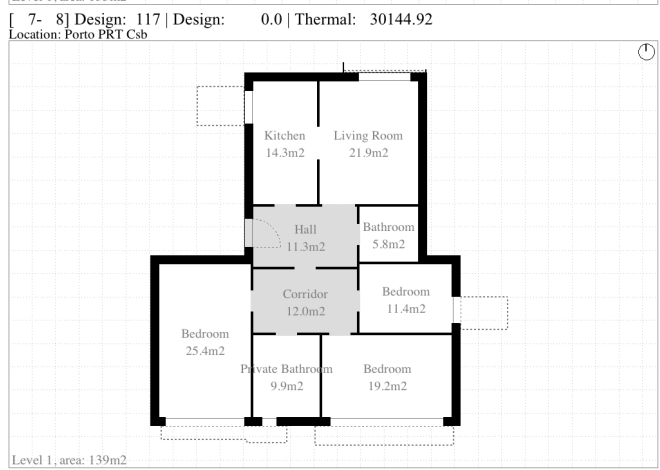
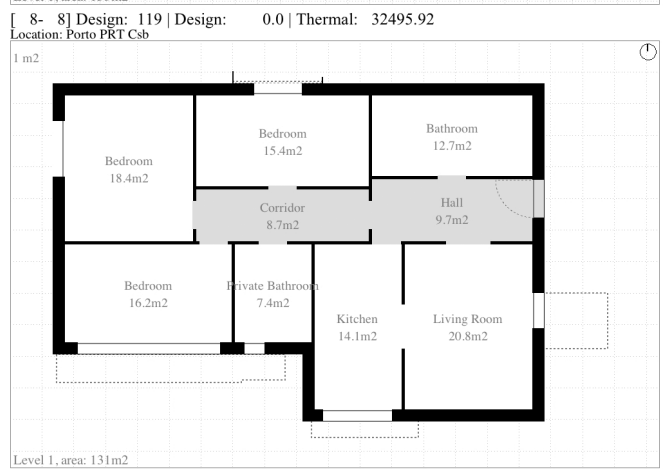

**Figure 1:** *Floor plan designs sorted by their thermal performance (reference scenario).*

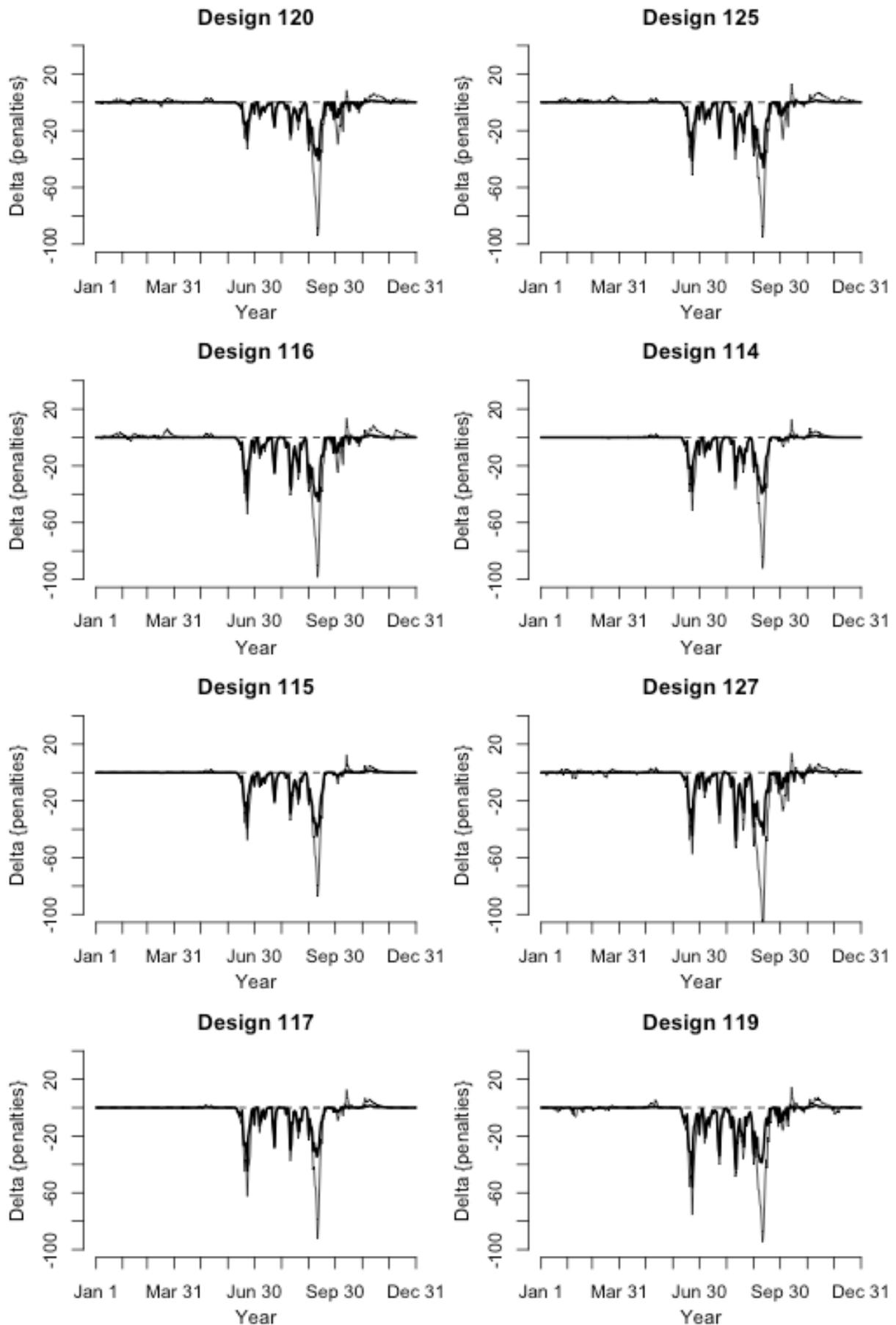

**Figure 2:** *Relative penalties difference between reference scenario and scenario B (strong line) and C (thin line). Dashed line is the reference scenario with no ventilation.*

**Acknowledgments**

The presented work is framed under the *Energy for Sustainability Initiative* of the University of Coimbra (UC) and has been partially supported by *Energy and Mobility for Sustainable Region Project* (EMSURE) CENTRO-07-0224-FEDER-002004 and project *Automatic Generation of Architectural Floor Plans with Energy Optimization* (GerAPlanO) CENTRO-07-0402-FEDER-038922.

**References**

Carrilho da Graça, G., Martins, N. R., & Horta, C. S. (2012). *Thermal and airflow simulation of a naturally ventilated shopping mall*. Energy and Buildings, *50*, 177–188. doi:10.1016/j.enbuild.2012.03.037

Faggianelli, G. A., Brun, A., Wurtz, E., & Muselli, M. (2014). *Natural cross ventilation in buildings on Mediterranean coastal zones*. Energy and Buildings, *77*, 206–218. doi:10.1016/j.enbuild.2014.03.042

Firląg, S., & Zawada, B. (2013). *Impacts of airflows, internal heat and moisture gains on accuracy of modeling energy consumption and indoor parameters in passive building*. Energy and Buildings, *64*, 372–383. doi:10.1016/j.enbuild.2013.04.024

Huang, C.-H., & Lin, P.-Y. (2014). *Influence of spatial layout on airflow field and particle distribution on the workspace of a factory*. Building and Environment, *71*, 212–222. doi:10.1016/j.buildenv.2013.09.014

Kao, H.-M., Chang, T.-J., Hsieh, Y.-F., Wang, C.-H., & Hsieh, C.-I. (2009). *Comparison of airflow and particulate matter transport in multi-room buildings for different natural ventilation patterns*. Energy and Buildings, *41*(9), 966–974. doi:10.1016/j.enbuild.2009.04.005

Karava, P., Stathopoulos, T., & Athienitis, a. K. (2011). *Airflow assessment in cross-ventilated buildings with operable façade elements*. Building and Environment, *46*(1), 266–279. doi:10.1016/j.buildenv.2010.07.022

Ng, L. C., Musser, A., Persily, A. K., & Emmerich, S. J. (2013). *Multizone airflow models for calculating infiltration rates in commercial reference buildings*. Energy and Buildings, *58*, 11–18. doi:10.1016/j.enbuild.2012.11.035

Rodrigues, E., Gaspar, A. R., & Gomes, Á. (2013a). *An approach to the multi-level space allocation problem in architecture using a hybrid evolutionary technique*. Automation in Construction, *35*, 482–498. doi:10.1016/j.autcon.2013.06.005

Rodrigues, E., Gaspar, A. R., & Gomes, Á. (2013b). *An evolutionary strategy enhanced with a local search technique for the space allocation problem in architecture, Part 1: Methodology*. Computer-Aided Design, *45*(5), 887–897. doi:10.1016/j.cad.2013.01.001

Rodrigues, E., Gaspar, A. R., & Gomes, Á. (2013c). *An evolutionary strategy enhanced with a local search technique for the space allocation problem in architecture, Part 2: Validation and performance tests*. Computer-Aided Design, *45*(5), 898–910. doi:10.1016/j.cad.2013.01.003

Rodrigues, E., Gaspar, A. R., & Gomes, Á. (2014a). *Automated approach for design generation and thermal assessment of alternative floor plans*. Energy and Buildings, *81*, 170–181. doi:10.1016/j.enbuild.2014.06.016

Rodrigues, E., Gaspar, A. R., & Gomes, Á. (2014b). *Improving thermal performance of floor plan designs using a sequential design variables optimization procedure*. Applied Energy, 132, 200-215. doi:10.1016/j.apenergy.2014.06.068

Stavrakakis, G. M., Zervas, P. L., Sarimveis, H., & Markatos, N. C. (2012). *Optimization of window-openings design for thermal comfort in naturally ventilated buildings*. Applied Mathematical Modelling, *36*(1), 193–211. doi:10.1016/j.apm.2011.05.052